\documentclass[journal]{IEEEtran}

\usepackage{cite}
\usepackage{amsmath,amssymb,amsfonts}
\usepackage{algorithmic}
\usepackage{graphicx}
\usepackage{textcomp}
\usepackage{xcolor}
\usepackage{hyperref}
\usepackage{url}
\usepackage{booktabs}

\hypersetup{
    colorlinks=true,
    linkcolor=blue,
    urlcolor=blue,
    citecolor=blue
}

\renewcommand{\footnoterule}{%
    \kern -3pt
    \hrule width \columnwidth height 0.4pt
    \kern 2pt
}

\def\BibTeX{{\rm B\kern-.05em{\sc i\kern-.025em b}\kern-.08em
    T\kern-.1667em\lower.7ex\hbox{E}\kern-.125emX}}

\begin{document}

\title{From Depth to Safe Depth: Uncertainty-Aware Monocular Perception for Mobile Robot Navigation}

\author{Saeid~Firouzi~Daghigh
        and~Saeed~Ayat%
\thanks{\textit{(Corresponding author: Saeid Firouzi Daghigh.)}}%
\thanks{The authors are with the Department of Computer Engineering and
Information Technology, Payame Noor University, Tehran, Iran (e-mail:
saeedmr881@gmail.com; dr.ayat@pnu.ac.ir).}}

\maketitle

\begin{abstract}
Monocular depth foundation models make it tempting to replace depth sensors on mobile robots with a single RGB camera. We show that, used off the shelf, this is unsafe: in a simulated indoor navigation benchmark, a standard Nav2 stack driven by Depth Anything V2 (metric) collides in 63\% of 120 episodes, because the model systematically over-estimates the distance to nearby obstacles. We propose a lightweight plug-in module that turns predicted depth into \emph{safe depth}. A 0.58\,M-parameter head on top of the \emph{frozen} depth model predicts, per pixel, a log-depth correction $\mu$ and an uncertainty $\sigma$, trained with a $\beta$-weighted Gaussian negative log-likelihood. The navigation stack consumes the one-sided bound $\hat d_{\text{safe}}=d\,e^{\mu-k\sigma}$, where $k$ is set by conformal calibration on a held-out world, and uncertainty also decides which pixels may be treated as floor. The correction halves the depth error (AbsRel $0.262\rightarrow0.134$) and reduces near-range over-estimation from 59.8\% to 12.0\%. In navigation, safe depth raises the success rate from 37\% to 70\% and lowers collisions from 63\% to 3\%, close to the 2\% of ground-truth depth. A constant safety margin calibrated to the same pixel coverage reaches only 30\% success because it freezes the robot; uncertainty-aware safety wins 52 of 56 discordant paired episodes against it. Gains hold on glass, texture-less walls and dark regions. Source code, pre-trained models, and test data results are publicly
available at: \url{https://github.com/saeed5959/depth_safe_robot}
\end{abstract}


\section{Introduction}

\IEEEPARstart{D}{epth} perception is the basis of obstacle avoidance for mobile robots. LiDAR and RGB-D sensors provide it directly, but add cost, weight and power, which is why camera-only navigation has long been attractive for small or low-cost platforms~\cite{ghasemi2023uav,simon2023mononav}. Monocular depth estimation has progressed from early multi-scale networks~\cite{eigen2014depth} through training on mixed datasets~\cite{ranftl2022midas} to foundation models such as Depth Anything V2~\cite{yang2024dav2} and Metric3D v2~\cite{hu2024metric3d}, which predict metric depth zero-shot in unseen scenes~\cite{liu2026survey}. Navigation systems increasingly build on such models, from monocular reconstruction for micro aerial vehicles~\cite{simon2023mononav} and risk-aware model predictive control~\cite{sharma2025monompc} to off-road stacks~\cite{marsal2026offroad} and safety layers for learned navigation policies~\cite{kim2025care,sridhar2024nomad}.

Foundation-model depth, however, is not equally reliable everywhere. Errors concentrate on texture-less surfaces, transparent objects, dark regions and, importantly, scenes outside the training distribution. For navigation the \emph{direction} of the error matters: under-estimating depth makes a robot cautious, over-estimating it makes obstacles appear farther than they are and leads to collisions. A cost map built directly from predicted depth treats every pixel as equally trustworthy, which MonoMPC~\cite{sharma2025monompc} identified as a main cause of collisions of standard Nav2 pipelines on Depth Anything point clouds.

Uncertainty quantification offers a principled way out. Aleatoric uncertainty can be learned by predicting a variance with a likelihood loss~\cite{kendall2017uncertainties}; evidential regression~\cite{amini2020evidential} and post-hoc heads on frozen networks such as BayesCap~\cite{upadhyay2022bayescap} avoid sampling at test time. A recent study combined several uncertainty methods with Depth Anything V2 and concluded that likelihood-based heads give useful pixel-wise uncertainty at little cost~\cite{landgraf2025critical}. What is less studied is how such uncertainty should enter a classical navigation stack, and whether it actually improves closed-loop safety compared with simply being more conservative everywhere.

This paper addresses that question with a deliberately simple design. Our contributions are:
\begin{itemize}
\item A plug-in \emph{safe-depth} module: a small head on a frozen depth foundation model that predicts a per-pixel bias correction $\mu$ and uncertainty $\sigma$, and outputs the one-sided bound $d\,e^{\mu-k\sigma}$ for an unchanged Nav2 stack. $k$ is set by conformal calibration on a held-out world.
\item An analysis of how depth uncertainty interacts with a downward-looking camera: shrinking depth along rays lifts the floor into phantom obstacles. We resolve this with an uncertainty-aware ground test.
\item A paired closed-loop evaluation with 120 episodes per method in procedurally generated indoor worlds with glass, blank walls and dark regions, comparing ground-truth depth, raw monocular depth, a constant margin with the same calibrated coverage, the corrected depth alone, and the full method, with McNemar tests on paired outcomes.
\end{itemize}

\section{Related Work}

\textbf{Monocular depth estimation.} Eigen et al.~\cite{eigen2014depth} introduced deep networks for single-image depth. MiDaS~\cite{ranftl2022midas} showed that mixing diverse datasets yields robust zero-shot transfer, but predicts relative depth. Depth Anything V2~\cite{yang2024dav2} trains a large teacher on synthetic data and distils it through pseudo-labelled real images, and provides metric variants fine-tuned on indoor (Hypersim) and outdoor data. Metric3D v2~\cite{hu2024metric3d} resolves scale ambiguity with a canonical camera space. A recent survey~\cite{liu2026survey} summarises this progress and notes that robustness and reliability remain open problems.

\textbf{Monocular navigation.} Camera-only navigation has been approached with learned traversability classifiers and generated depth~\cite{ghasemi2023uav}, with monocular metric reconstruction feeding a motion-primitive planner~\cite{simon2023mononav}, and with end-to-end diffusion policies~\cite{sridhar2024nomad}. MonoMPC~\cite{sharma2025monompc} conditions a learned probabilistic collision model on the estimated point cloud and plans with risk-aware MPC. CARE~\cite{kim2025care} adds a repulsive-force safety layer to visual navigation foundation models using monocular depth. Marsal et al.~\cite{marsal2026offroad} compare LiDAR with monocular depth inside an open-source off-road stack. Unlike these works, we keep the standard planner and cost map, and change only the depth that is fed to them.

\textbf{Uncertainty in depth.} Kendall and Gal~\cite{kendall2017uncertainties} distinguished aleatoric and epistemic uncertainty and learned heteroscedastic variance for depth regression. Seitzer et al.~\cite{seitzer2022pitfalls} showed that the Gaussian NLL lets networks ``give up'' on hard samples and proposed the $\beta$-NLL. Evidential regression~\cite{amini2020evidential} predicts both uncertainty types in one pass. BayesCap~\cite{upadhyay2022bayescap} learns calibrated uncertainty for frozen networks, the setting we adopt. Landgraf et al.~\cite{landgraf2025critical} benchmarked uncertainty methods on Depth Anything V2. Conformal prediction~\cite{angelopoulos2023conformal} provides distribution-free coverage guarantees from a calibration set; we use it to set the safety factor $k$.

\section{Method}

\subsection{Overview}
\begin{figure*}[t]
\centering
\includegraphics[width=\textwidth]{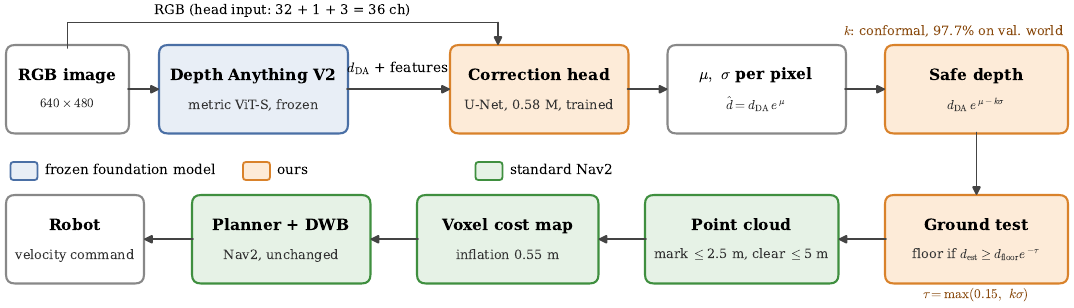}
\caption{Safe-depth pipeline. A frozen Depth Anything V2 model predicts metric depth $d_{\text{DA}}$. A small head, fed with the decoder features, $\log d_{\text{DA}}$ and the RGB image, predicts a per-pixel log-depth correction $\mu$ and uncertainty $\sigma$. The lower bound $d_{\text{DA}}e^{\mu-k\sigma}$, with $k$ set by conformal calibration on a validation world, passes through an uncertainty-aware ground test before an unmodified Nav2 stack builds its cost map. Only the orange components are ours; the blue model and green navigation stack are unchanged.}
\label{fig:pipeline}
\end{figure*}

Fig.~\ref{fig:pipeline} summarises the pipeline. An RGB frame is passed through a frozen metric Depth Anything V2 model (ViT-S, Hypersim checkpoint, input 518\,px). Its depth $d_{\text{DA}}$, decoder features and the image feed a small head that outputs $\mu$ and $s=\log\sigma$ per pixel. The resulting depth image is converted to a point cloud and consumed by an unmodified Nav2~\cite{macenski2020marathon} voxel cost map, planner and controller.

\subsection{Correction and uncertainty head}
We hook the 32-channel feature map that enters the final convolution of the DPT decoder. Together with $\log d_{\text{DA}}$ and the normalised RGB image it forms a 36-channel input at half the network resolution. The head is a small U-Net: two convolution blocks, two stride-2 stages with residual dilated convolutions (dilations up to 8, for context on large uniform surfaces), two up-sampling stages with skip connections, and a $1\times1$ output layer, 0.58\,M parameters in total. The output layer is initialised to $\mu=0$, $\sigma=0.2$, so training starts exactly from the raw model. Working in log-depth makes errors multiplicative, which matches how monocular depth errors scale with distance. The corrected depth and its uncertainty are
\begin{equation}
\log\hat d = \log d_{\text{DA}} + \mu,\qquad \sigma=e^{s}.
\end{equation}

\subsection{Training objective}
With the residual $r=\log d_{\text{GT}}-\log\hat d$, the Gaussian negative log-likelihood is
\begin{equation}
\mathcal{L}_{\text{NLL}} = \tfrac12\, r^2 e^{-2s} + s .
\end{equation}
Its minimum over $s$ lies at $\sigma=|r|$, so the head learns, without any uncertainty labels, the typical error size of each kind of pixel, while $\mu$ learns the systematic part of the error. Because the gradient of $\mu$ is scaled by $\sigma^{-2}$, plain NLL tends to inflate $\sigma$ on hard pixels instead of correcting them. We therefore use the $\beta$-NLL~\cite{seitzer2022pitfalls}, multiplying the loss by $\mathrm{sg}(\sigma^{2\beta})$ with $\beta=0.5$, where $\mathrm{sg}$ stops the gradient. Pixels are weighted by $w=1/\max(d_{\text{GT}},0.5\,\text{m})$, since nearby obstacles cause collisions, and pixels with invalid ground truth or depth outside $[0.1,10]$\,m are masked.

\subsection{Safe depth and calibration}
The depth passed to navigation is the one-sided lower bound
\begin{equation}
\hat d_{\text{safe}} = d_{\text{DA}}\, e^{\mu-k\sigma}.
\end{equation}
Instead of fixing $k=2$, which assumes a perfectly calibrated Gaussian, we choose $k$ by split-conformal calibration~\cite{angelopoulos2023conformal}: on a held-out validation world we compute the normalised scores $z=(\log\hat d-\log d_{\text{GT}})/\sigma$ of pixels closer than 3\,m and set $k$ to their 97.7\% quantile, so that $d_{\text{GT}}\ge\hat d_{\text{safe}}$ holds for 97.7\% of near pixels on that world. This gave $k=3.0$. For the constant-margin baseline $d_{\text{DA}}e^{-c}$, $c$ is calibrated the same way to the same coverage, which gave $c=0.73$ (depth scaled by 0.48). Both baselines are therefore equally safe at the pixel level; they differ only in \emph{where} the margin is spent.

\subsection{Integrating safe depth with the cost map}
\textbf{Phantom floor.} The camera looks down at the floor. Shrinking depth along a ray moves a floor point towards the camera \emph{and upwards}: for a camera at height $H$, a floor pixel whose depth is under-estimated by a fraction $\epsilon$ is lifted by $H\epsilon$ at any range. With $H=0.55$\,m, a 14\% under-estimate already exceeds the cost map's obstacle threshold, and safe depth turned the whole floor into lethal cells. We therefore classify ground pixels explicitly. Each pixel's ray meets the floor plane at depth $d_{\text{floor}}$, obtained from the camera intrinsics and its pose above the floor. A pixel is ground if
\begin{equation}
\log d_{\text{est}}-\log d_{\text{floor}} \ge -\tau,
\end{equation}
where $d_{\text{est}}$ is the method's best estimate ($d_{\text{DA}}$ for the raw and margin variants, $\hat d$ otherwise), and its depth is replaced by $d_{\text{floor}}$, so it clears but never marks the cost map. We use $\tau=0.15$ for all variants and, for the safe variant, $\tau=\max(0.15,\,k\sigma)$: a pixel is treated as floor if the floor lies within its own uncertainty band. Uncertainty thus decides both \emph{where} obstacles are and \emph{whether} a pixel is an obstacle at all.

\textbf{Range and timing.} Monocular depth is reliable at short range, so for all monocular variants obstacles are marked within 2.5\,m and cleared by ray tracing up to 5\,m (4\,m and 5\,m for ground truth). Because inference finishes about one second after the image was captured, the depth node republishes the camera intrinsics with the depth image's time stamp; otherwise the point-cloud conversion silently drops most frames.

\section{Experimental Setup}

\subsection{Simulation and robot}
Experiments run in Gazebo Fortress with ROS~2 Humble. The robot is a differential-drive base ($0.40\times0.30$\,m, maximum speed 0.3\,m/s) with a 640$\times$480 RGB-D camera (60$^\circ$ horizontal field of view) mounted 0.55\,m above the floor and pitched down by 0.2\,rad. The depth channel of the same camera provides ground truth. To isolate perception, all methods use ground-truth odometry, the same Nav2 configuration (NavFn planner, DWB controller, voxel cost maps, robot radius 0.25\,m, inflation 0.55\,m), and identical simulation settings.

\subsection{Worlds}
\begin{figure}[t]
\centering
\includegraphics[width=\columnwidth]{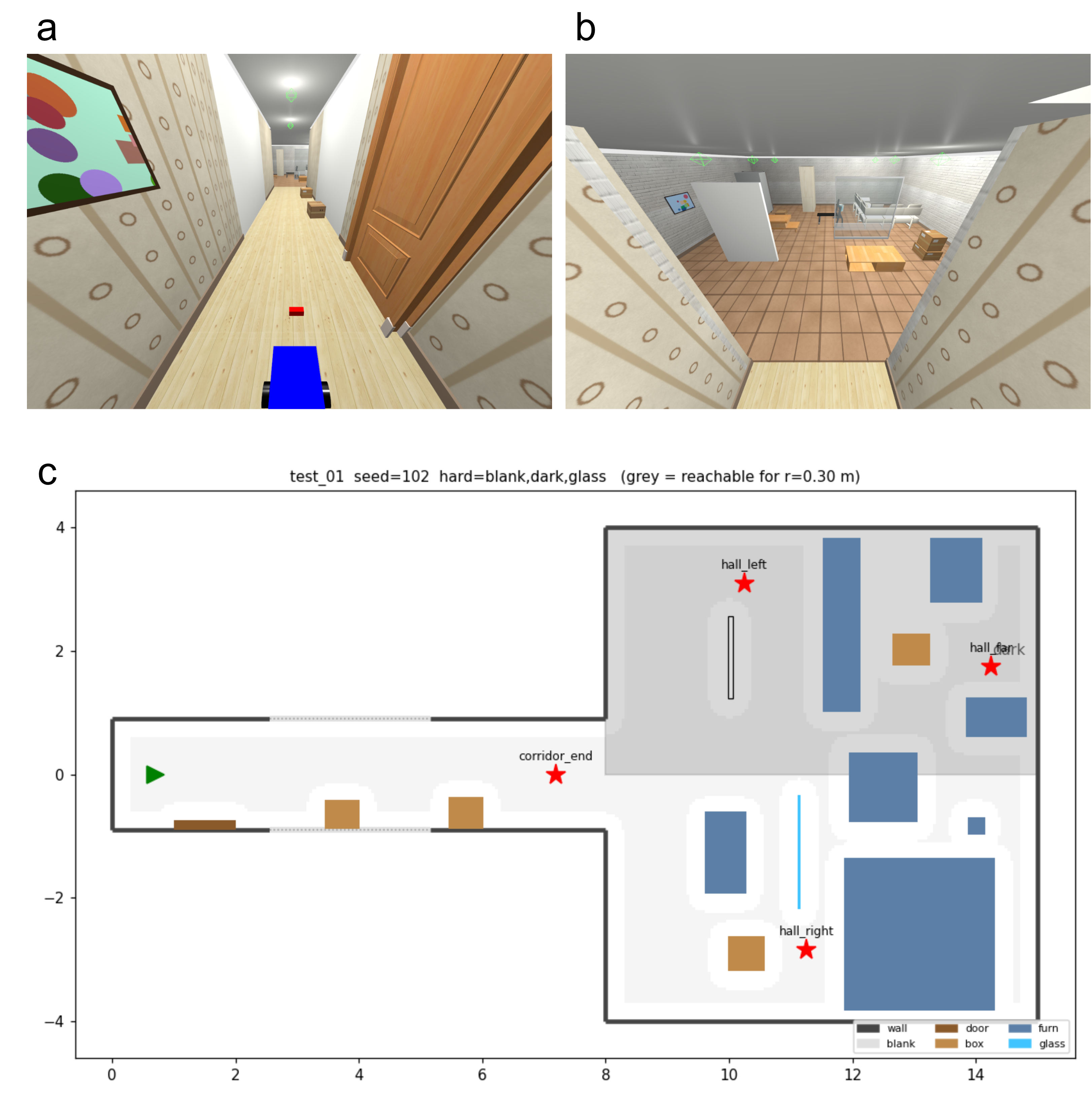}
\caption{Simulation environment. (a) Corridor seen from behind the robot. (b) View from the corridor into the hall with furniture, boxes and the glass partition. (c) Top view of test world \texttt{test\_01} with walls, blank panels, boxes, furniture, the glass partition, the dark region and the four canonical goals; grey marks the area reachable by a robot of radius 0.30\,m.}
\label{fig:world}
\end{figure}

A procedural generator builds indoor worlds from a seed: an 8\,m corridor (1.8\,m wide) with doors, pictures and cardboard boxes, opening into a $7\times8$\,m hall with furniture from the AWS RoboMaker small-house set (Fig.~\ref{fig:world}). Three hard cases can be enabled: a free-standing \emph{glass} partition (tinted pane with a frosted safety band and rails), \emph{blank} texture-less white walls, and a \emph{dark} half of the hall lit at about 10\%. Five training worlds use different subsets of hard cases; one of them (\texttt{train\_03}) is held out for validation and calibration. Four test worlds are never used for training or tuning: \texttt{test\_clean} without hard cases, and \texttt{test\_01}--\texttt{03} with all three.

\subsection{Episodes and metrics}
\begin{figure}[t]
\centering
\includegraphics[width=\columnwidth]{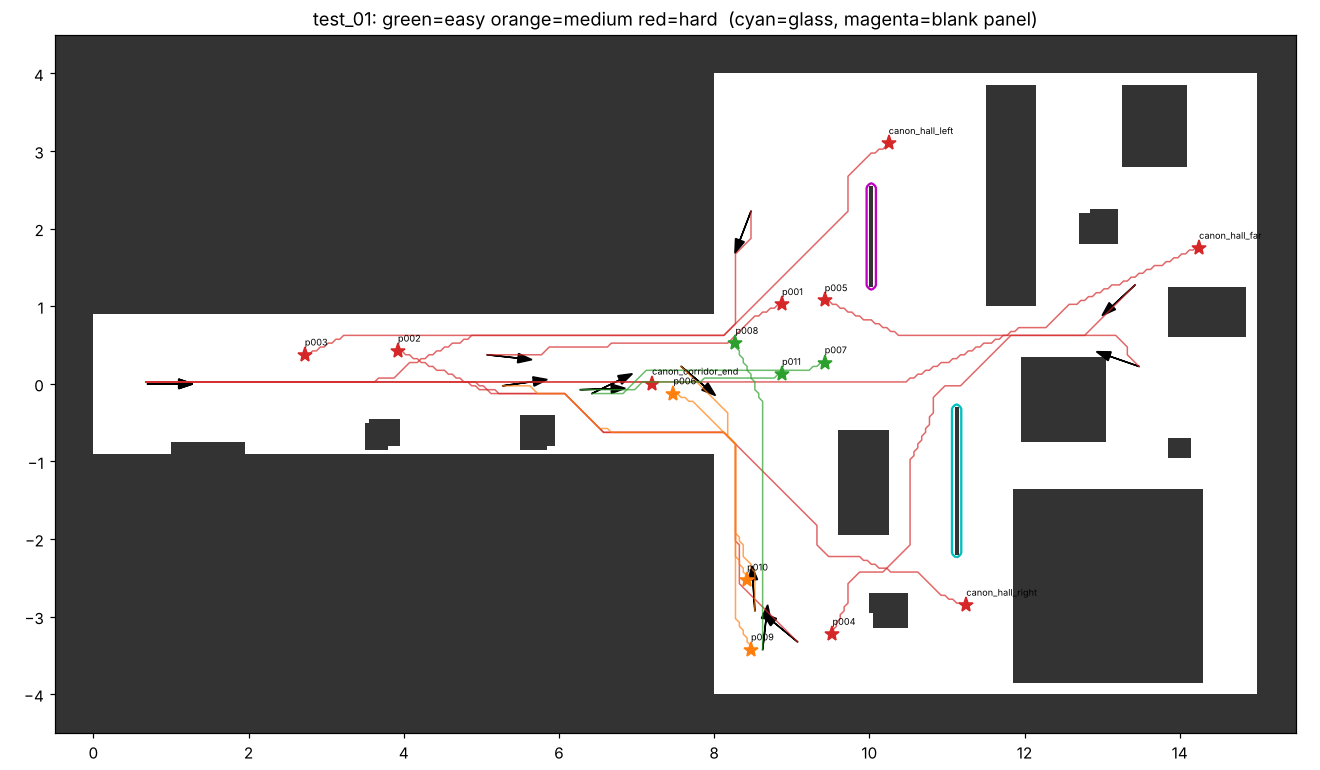}
\caption{Start poses (arrows), goals (stars) and A* reference paths of the sampled episodes in \texttt{test\_01}, coloured by difficulty: easy (green), medium (orange), hard (red). Cyan and magenta outline the glass partition and the blank panel.}
\label{fig:episodes}
\end{figure}

Each test world has 30 episodes: the four canonical goals from the spawn pose and 11 sampled start-goal pairs, each run twice with small start perturbations ($\pm0.1$\,m, $\pm10^\circ$), giving 120 paired episodes per method (Fig.~\ref{fig:episodes}). Starts and goals keep at least 0.45\,m clearance and are connected by an A* path for a 0.30\,m robot. An episode is labelled \emph{easy} with line of sight, \emph{hard} if its path passes a hard case or a narrow gap, and \emph{medium} otherwise. It ends with \emph{success} when Nav2 reaches the goal within 0.3\,m, \emph{collision} on any contact of the robot body (bumper sensor), \emph{timeout} after $2.5\,L_{A^*}/v_{\max}+20$\,s, \emph{stuck} after 40\,s without 0.1\,m progress, or \emph{failure} when Nav2 aborts. We report success rate with Wilson 95\% intervals, outcome rates, SPL, time to goal and minimum clearance. Because all methods run the same episodes, we compare them with exact McNemar tests on the discordant pairs.

\subsection{Training data}
RGB images and ground-truth depth were recorded on the five training worlds while the robot executed episodes with both ground-truth and raw monocular depth, so that the data include the close approaches that precede monocular collisions. A frame was stored every 0.2\,m or 10$^\circ$ of motion. Training uses 7{,}045 frames from four worlds; \texttt{train\_03} is used for validation and calibration. The head was trained for 5 epochs with AdamW (learning rate $10^{-3}$, one-cycle schedule), batch size 6, horizontal flips and brightness augmentation, on a single GTX~1650 (4\,GB) with the depth model frozen in half precision.

\subsection{Compared methods}
\textbf{GT:} ground-truth depth (upper bound). \textbf{Raw:} $d_{\text{DA}}$. \textbf{Margin:} $d_{\text{DA}}e^{-c}$, $c=0.73$. \textbf{$\mu$:} corrected depth $\hat d$. \textbf{Safe:} $\hat d_{\text{safe}}$ with $k=3.0$. All monocular variants share the same depth node, ground test and cost-map settings.

\section{Results}

\subsection{Depth accuracy}
Table~\ref{tab:depth} reports depth accuracy on the test worlds. Raw Depth Anything V2 is far from its benchmark accuracy in this setting: the mean relative error is 26\%, only 62.5\% of pixels lie within 25\% of the truth, and 59.8\% of pixels closer than 3\,m are over-estimated by more than 0.2\,m. The error is not a constant scale factor, as the ratio of predicted to true depth varied between roughly 0.8 and 2 across scenes. The head halves AbsRel, raises $\delta_{1.25}$ to 91.9\% and cuts dangerous near-range over-estimation five-fold. On the validation world the area under the sparsification error curve was 0.028, indicating that $\sigma$ ranks errors well.

\begin{table}[t]
\centering
\caption{Depth accuracy on the test worlds. Near over-estimation: fraction of pixels closer than 3\,m whose prediction exceeds the truth by more than 0.2\,m.}
\label{tab:depth}
\begin{tabular}{lcccc}
\toprule
Method & AbsRel $\downarrow$ & $\delta_{1.25}\uparrow$ & Near over-est.\ $\downarrow$ & Mean $\sigma$ \\
\midrule
DA2 raw & 0.262 & 62.5\% & 59.8\% & - \\
DA2 + head ($\mu$) & \textbf{0.134} & \textbf{91.9\%} & \textbf{12.0\%} & 0.131 \\
\bottomrule
\end{tabular}
\end{table}

\subsection{Navigation}
Table~\ref{tab:nav} summarises the 120 paired episodes. With ground-truth depth the stack succeeds in 93\% of episodes. Raw monocular depth collides in 63\%; it never times out or gets stuck, which is the signature of over-confidence: obstacles are placed too far away and the robot drives into them. Its minimum clearance of 0.25\,m is exactly the contact distance of the body.

The constant margin removes most collisions (5\%) but leaves the robot stuck or timed out in 65\% of episodes. Shrinking every depth by half narrows the 1.8\,m corridor and the gaps between furniture until no path remains. It is \emph{less} successful than raw depth. Correcting the bias ($\mu$) raises success to 60\% and lowers collisions to 15\%. Adding uncertainty gives the best monocular result: 70\% success with 3\% collisions, close to the 2\% of ground truth, and a minimum clearance of 0.45\,m. The safe variant takes slightly longer than ground truth (26\,s vs.\ 24.5\,s) because it keeps more distance where it is unsure.

\begin{table*}[t]
\centering
\caption{Closed-loop navigation on 120 paired test episodes. Success with Wilson 95\% interval; outcome rates; SPL; mean time of successful episodes; median minimum clearance.}
\label{tab:nav}
\begin{tabular}{lcccccc}
\toprule
Method & Success [95\% CI] $\uparrow$ & Collision $\downarrow$ & Stuck/timeout/fail $\downarrow$ & SPL $\uparrow$ & Time (s) & Min.\ clear.\ (m) \\
\midrule
GT depth & 0.93 [0.87, 0.96] & 0.02 & 0.05 & 0.92 & 24.5 & 0.49 \\
\midrule
Mono raw & 0.37 [0.29, 0.46] & 0.63 & 0.00 & 0.36 & 14.7 & 0.25 \\
Mono + margin ($c=0.73$) & 0.30 [0.23, 0.39] & 0.05 & 0.65 & 0.28 & 18.0 & 0.45 \\
Mono + $\mu$ & 0.60 [0.51, 0.68] & 0.15 & 0.25 & 0.55 & 22.0 & 0.38 \\
Mono + $\mu-k\sigma$ (ours, $k=3.0$) & \textbf{0.70} [0.61, 0.77] & \textbf{0.03} & 0.27 & \textbf{0.63} & 26.0 & 0.45 \\
\bottomrule
\end{tabular}
\end{table*}

Table~\ref{tab:mcnemar} shows that the differences are significant on paired outcomes. The safe variant wins 45 of 50 discordant episodes against raw depth and 52 of 56 against the margin. Adding $\sigma$ to the correction helps on its own (20 vs.\ 8, $p=0.036$). Ground truth remains clearly better (30 vs.\ 2).

\begin{table}[t]
\centering
\caption{Exact McNemar tests on paired episodes. $b$: first method succeeds and the second fails; $c$: the reverse.}
\label{tab:mcnemar}
\begin{tabular}{lccc}
\toprule
Comparison & $b$ & $c$ & $p$ \\
\midrule
Safe vs.\ raw & 45 & 5 & $4\times10^{-9}$ \\
Safe vs.\ margin & 52 & 4 & $1\times10^{-11}$ \\
$\mu$ vs.\ raw & 33 & 5 & $4\times10^{-6}$ \\
Safe vs.\ $\mu$ & 20 & 8 & 0.036 \\
Safe vs.\ GT & 2 & 30 & $2\times10^{-7}$ \\
\bottomrule
\end{tabular}
\end{table}

\subsection{Hard cases}
Table~\ref{tab:hard} breaks success down by the hard case an episode's path passes; an episode can count in several groups. The ordering raw $<\mu<$ safe holds in every group, the margin is never better than $\mu$, and the safe variant improves over raw depth by 33-37 percentage points in each group, including the scenes without hard cases. Glass remains the hardest case for all monocular variants: the safety band and rails provide only weak cues, and the glass groups are small (16 episodes), so their intervals are wide. Note that the simulated depth camera sees glass as solid, which makes the ground-truth baseline optimistic compared with real depth sensors.

\begin{table}[t]
\centering
\caption{Success rate by hard case on the path.}
\label{tab:hard}
\begin{tabular}{lccccc}
\toprule
Hard case & GT & Raw & Margin & $\mu$ & Safe \\
\midrule
Glass & 0.94 & 0.00 & 0.25 & 0.31 & \textbf{0.37} \\
Blank wall & 0.98 & 0.31 & 0.25 & 0.54 & \textbf{0.67} \\
Dark & 0.96 & 0.25 & 0.21 & 0.50 & \textbf{0.58} \\
None & 0.88 & 0.39 & 0.35 & 0.65 & \textbf{0.76} \\
\bottomrule
\end{tabular}
\end{table}

\section{Discussion and Limitations}

\textbf{Where the margin goes matters.} The margin and the safe variant are calibrated to the same pixel coverage, yet their outcomes differ by 40 percentage points. A uniform margin spends caution everywhere, including on well-textured walls and furniture that the model sees reliably. $\sigma$ spends it where errors actually occur. Bias correction alone is not sufficient either: $\mu$ still collides in 15\% of episodes, because its remaining errors are not systematic and only the uncertainty term can cover them.

\textbf{Uncertainty and geometry.} Lowering depth along camera rays has side effects that a cost map does not expect: it lifts the floor and pulls side walls inwards. The first we resolve with the uncertainty-aware ground test. The second remains: with large $k$, narrow passages shrink, and part of the remaining 27\% of non-collision failures are such conservative stops. A calibration target defined on navigation outcomes, or applying the bound only to nearby obstacles, could trade these off better than the pixel-level coverage used here.

\textbf{Limitations.} All experiments are in simulation with one robot, one camera configuration and one depth model (ViT-S). Ground-truth odometry isolates perception but hides localisation error. The ground test assumes a flat floor. The head is trained on simulated data from the same generator family as the test worlds, although on different layouts, so transfer to real images needs to be verified. Monocular inference at about 1\,Hz on a 4\,GB GPU also limits the speed at which the robot can safely drive.

\section{Conclusion}
Off-the-shelf monocular depth from a foundation model is not a safe drop-in replacement for a depth sensor: in our benchmark it made a standard navigation stack collide in most episodes. A small head on the frozen model, trained to predict a correction and an uncertainty, and a conformally calibrated lower bound turn this depth into safe depth that an unmodified Nav2 stack can use. It nearly doubles the success rate and brings collisions close to those of ground-truth depth, while a constant margin with the same calibrated safety level leaves the robot unable to move. Future work will validate the module on real robots, use navigation-aware calibration, and combine aleatoric with epistemic uncertainty for unseen environments.


\begin{thebibliography}{99}

\bibitem{ghasemi2023uav}
H.~Ghasemi, E.~Balazadeh, S.~Malekpour, A.~Kalhor, and M.~Tale~Masouleh, ``Autonomous unmanned aerial vehicle indoor navigation based on deep learning approaches,'' in \emph{Proc. Int. Conf. Robotics and Mechatronics (ICRoM)}, 2023.

\bibitem{simon2023mononav}
N.~Simon and A.~Majumdar, ``MonoNav: MAV navigation via monocular depth estimation and reconstruction,'' in \emph{Proc. Int. Symp. Experimental Robotics (ISER)}, 2023, arXiv:2311.14100.

\bibitem{sridhar2024nomad}
A.~Sridhar, D.~Shah, C.~Glossop, and S.~Levine, ``NoMaD: Goal masked diffusion policies for navigation and exploration,'' in \emph{Proc. IEEE Int. Conf. Robotics and Automation (ICRA)}, 2024, arXiv:2310.07896.

\bibitem{sharma2025monompc}
B.~Sharma, P.~Jadhav, P.~Paul, K.~M.~Krishna, and A.~K.~Singh, ``MonoMPC: Monocular vision based navigation with learned collision model and risk-aware model predictive control,'' 2025, arXiv preprint.

\bibitem{marsal2026offroad}
R.~Marsal, Q.~Picard, A.~Poir\'e, S.~Kerbourc'h, T.~Toralba, C.~Yver, A.~Chapoutot, and D.~Filliat, ``An open-source LiDAR and monocular off-road autonomous navigation stack,'' 2026, arXiv:2604.03096.

\bibitem{eigen2014depth}
D.~Eigen, C.~Puhrsch, and R.~Fergus, ``Depth map prediction from a single image using a multi-scale deep network,'' in \emph{Advances in Neural Information Processing Systems (NeurIPS)}, 2014.

\bibitem{ranftl2022midas}
R.~Ranftl, K.~Lasinger, D.~Hafner, K.~Schindler, and V.~Koltun, ``Towards robust monocular depth estimation: Mixing datasets for zero-shot cross-dataset transfer,'' \emph{IEEE Trans. Pattern Anal. Mach. Intell.}, vol.~44, no.~3, pp.~1623-1637, 2022.

\bibitem{yang2024dav2}
L.~Yang, B.~Kang, Z.~Huang, Z.~Zhao, X.~Xu, J.~Feng, and H.~Zhao, ``Depth Anything V2,'' in \emph{Advances in Neural Information Processing Systems (NeurIPS)}, 2024.

\bibitem{hu2024metric3d}
M.~Hu, W.~Yin, C.~Zhang, Z.~Cai, X.~Long, H.~Chen, K.~Wang, G.~Yu, C.~Shen, and S.~Shen, ``Metric3D v2: A versatile monocular geometric foundation model for zero-shot metric depth and surface normal estimation,'' \emph{IEEE Trans. Pattern Anal. Mach. Intell.}, 2024.

\bibitem{upadhyay2022bayescap}
U.~Upadhyay, S.~Karthik, Y.~Chen, M.~Mancini, and Z.~Akata, ``BayesCap: Bayesian identity cap for calibrated uncertainty in frozen neural networks,'' in \emph{Proc. Eur. Conf. Computer Vision (ECCV)}, 2022.

\bibitem{landgraf2025critical}
S.~Landgraf, R.~Qin, and M.~Ulrich, ``A critical synthesis of uncertainty quantification and foundation models in monocular depth estimation,'' 2025, arXiv:2501.08188.

\bibitem{kim2025care}
J.~Kim, J.~Sim, W.~Kim, K.~Sycara, and C.~Nam, ``Enhancing safety of foundation models for visual navigation through collision avoidance via repulsive estimation,'' 2025, arXiv:2506.03834.

\bibitem{liu2026survey}
M.~Liu, X.~Lyu, Y.-T.~Sun, Y.-H.~Huang, Z.~Yang, P.~Dai, and X.~Qi, ``Monocular depth estimation from a single image: Progress and opportunities,'' 2026, arXiv:2609.01172.

\bibitem{kendall2017uncertainties}
A.~Kendall and Y.~Gal, ``What uncertainties do we need in Bayesian deep learning for computer vision?'' in \emph{Advances in Neural Information Processing Systems (NeurIPS)}, 2017.

\bibitem{amini2020evidential}
A.~Amini, W.~Schwarting, A.~Soleimany, and D.~Rus, ``Deep evidential regression,'' in \emph{Advances in Neural Information Processing Systems (NeurIPS)}, 2020.

\bibitem{seitzer2022pitfalls}
M.~Seitzer, A.~Tavakoli, D.~Antic, and G.~Martius, ``On the pitfalls of heteroscedastic uncertainty estimation with probabilistic neural networks,'' in \emph{Proc. Int. Conf. Learning Representations (ICLR)}, 2022.

\bibitem{angelopoulos2023conformal}
A.~N.~Angelopoulos and S.~Bates, ``Conformal prediction: A gentle introduction,'' \emph{Found. Trends Mach. Learn.}, vol.~16, no.~4, pp.~494-591, 2023.

\bibitem{macenski2020marathon}
S.~Macenski, F.~Mart\'in, R.~White, and J.~Gin\'es~Clavero, ``The Marathon 2: A navigation system,'' in \emph{Proc. IEEE/RSJ Int. Conf. Intelligent Robots and Systems (IROS)}, 2020.

\end{thebibliography}
\end{document}